# Left Ventricle Contouring in Cardiac Images Based on Deep Reinforcement Learning

Sixing Yin, Yameng Han, and Shufang Li

School of Information and Communication Engineering

Beijing University of Posts and Telecommunications, China

Abstract—Medical image segmentation is one of the important tasks of computer-aided diagnosis in medical image analysis. Since most medical images have the characteristics of blurred boundaries and uneven intensity distribution, through existing segmentation methods, the discontinuity within the target area and the discontinuity of the target boundary are likely to lead to rough or even erroneous boundary delineation. In this paper, we propose a new iterative refined interactive segmentation method for medical images based on agent reinforcement learning, which focuses on the problem of target segmentation boundaries. We model the dynamic process of drawing the target contour in a certain order as a Markov Decision Process (MDP) based on a deep reinforcement learning method. In the dynamic process of continuous interaction between the agent and the image, the agent tracks the boundary point by point in order within a limited length range until the contour of the target is completely drawn. In this process, the agent can quickly improve the segmentation performance by exploring an interactive policy in the image. The method we proposed is simple and effective. At the same time, we evaluate our method on the cardiac MRI scan data set. Experimental results show that our method has a better segmentation effect on the left ventricle in a small number of medical image data sets, especially in terms of segmentation boundaries, this method is better than existing methods. Based on our proposed method, the dynamic generation process of the predicted contour trajectory of the left ventricle will be displayed online at https://github.com/H1997ym/LV-contour-trajectory.

## I. INTRODUCTION

In recent years, with the rapid development and popularization of medical imaging equipment, medical imaging data analysis has become an important auxiliary diagnosis and treatment method in the medical field. Imaging technology has become a non-invasive means to observe the anatomical structure of organs and tissues. For example, computer tomography(CT), magnetic resonance imaging (MRI), positron emission tomography (PET) and other technologies can objectively and clearly reflect the tissue structure, pathological changes, and provide important help for doctors from location forensics to guided treatment, and a strong guarantee for clinical diagnosis and biomedical research[1]. Therefore, relevant experts have always attached great importance to medical image processing technology, and medical image segmentation is an important step in image processing technology. It is not only the primary problem to be solved for image analysis and recognition, but also a bottleneck restricting the development and application of related technologies such as visualization, image registration and fusion of different modalities, and 3D reconstruction in medical image processing[2]. This paper discusses the application of deep learning in medical image processing from the perspective of medical image segmentation.

Medical image segmentation is the extraction of specific anatomical organs or lesions in medical images. It is a particularly important processing step for automatic image pattern recognition and scene analysis and understanding. It is also a bridge from low-level image processing to high-level image understanding. Medical image segmentation has important academic research significance and application value in research and practice fields such as medical research, clinical diagnosis, pathological analysis, and image information processing. It is embodied in the following aspects:

(1) Extract quantitative information of special tissues in the image, such as calculating the volume of human organs, tissues or lesions, so as to perform quantitative analysis on medical images.

(2) The subsequent processing of images facilitates image analysis and understanding, such as image registration, fusion, and recognition of different imaging devices. These technologies have greatly expanded the scope of medical images used by clinicians.

(3) It is used in clinical medical application systems to retrieve medical images, such as 3D reconstruction of medical images, pathological research, anatomical references, and the formulation of surgical procedures.

Due to the interference of various complex factors inside the human body and medical imaging equipment, as well as the influence of imaging noise, artifacts and partial volume effect, the actual medical images obtained have the characteristics of blur, uneven grayscale distribution, and defects in the boundaries between different tissues. In addition, the anatomical structure of the human body has a certain degree of complexity and individuals have great differences in their response to pathology or lesions under their respective physiological conditions. These complicated conditions have brought great difficulty to medical image segmentation. At the same time, in the actual medical field application, the vast majority of cases will face the situation of less target data sources and small scale. In the absence of large-scale training data sets, it is particularly important to design a reasonable network architecture to adapt to the reality of small-scale data sets. This is an important link in the combination of technology and actual scenarios.

Although the segmentation technology based on deep convolutional network has been widely used in various medical imaging modes, and has shown a broad future, it has overcome the limitations of conventional segmentation technology. At present, a lot of work has been done, normally using an end-to-end dense predictor to classify each pixel in the image to directly predict the label of the pixel. Neural networks focus on extracting features from specific regions[3], but due to the complex anatomical structure of the human body, there may be defects within the tissues, and there are often blurred boundaries between different tissues. Therefore, the convolutional neural network may easily ignore the local spatial information where the contour boundary between the target object and the background is located, which makes it difficult to classify the pixels near the target boundary, resulting in an incorrect contour segmentation. In recent years, some researchers have improved the segmentation boundary problem by designing more complex loss functions[4], adding additional contour edge detectors[5], and using some conventional methods such as conditional random fields (CRF)[22] or manual features. However, these methods often require more complex designs, which reduces the efficiency of segmentation.

In order to solve the above problems, we propose an interactive image segmentation method based on deep reinforcement learning technology. This method uses a small amount of training set to make the image segmentation boundary achieve better results, and the method is simple and effective. We model the iterative dynamic process of the agent interacting with the image as a Markov Decision Process (MDP). By combining the rewards obtained, the agent tracks the contour boundary of the target point by point within a limited length range until the target contour is completely drawn. In our method, we regard the agent as a paintbrush that draws the contour of the target, and its drop position for each interaction is centered on its current position, and drops down at one of the eight directions surrounding it. The agent selects an appropriate initial drop position, and then effectively interacts with the image, obtains corresponding feedback according to the predefined segmentation metric, and can explore a better segmentation policy through the backpropagation method to refine the segmentation result. In the training process, we divide the segmentation task into two sub-tasks, one is to find the starting point of the contour trajectory, and the other is to ignore the starting point and is responsible for the generation of the subsequent contour trajectory. In the second subtask, for each training image, we can get a lot of contour trajectories with different initial starting points. Therefore, even if we only have a small number of sample images, we can still get a training set with a large amount of trajectory data. Based on this, in the training process of the trajectory model, we can extract richer and more detailed local space information where the target segmentation boundary is located, so as to achieve a better segmentation effect.

We evaluate our method on the data set of cardiac MRI scans. The experimental results show that our method uses a smaller amount of training set to produce better results for the image segmentation boundary. The contributions of this paper are as follows:

1) We propose an interactive image segmentation method based on deep reinforcement learning, which imitates the process of drawing the contour boundary of the target point by point.

2) We use a smaller amount of sample images, but obtain a large number of training sets of trajectory data, so that we can extract richer and more detailed image features to achieve better segmentation results.

3) We use the relative rewards obtained in the continuous interactive dynamic process between the agent and the image, and we can adjust the policy in time, refine the segmentation, and improve the speed and efficiency of the segmentation.

## II Related Work

In this section, the conventional image segmentation method based on convolutional neural network and the content related to this work will be briefly reviewed in the following context.

### A. Conventional segmentation methods

Conventional image segmentation is to divide the image into several disjoint areas according to the characteristics of grayscale, color, spatial texture, etc., so that these characteristics show consistency or similarity in the same area, but show obviously different between different areas. Conventional segmentation technologies include edge-based image segmentation technology, threshold-based image segmentation technology, region-based image segmentation technology, etc.[6].

Threshold segmentation method is one of the commonly used segmentation techniques, and its essence is to automatically determine the optimal threshold value according to certain standards, and use these pixels according to the gray level to achieve clustering. The region-based segmentation method is a segmentation technique based on directly finding new regions, which can be divided into two basic methods: region growth and region splitting and merging. The segmentation method based on edge detection splits the image by detecting the edges of different regions. It uses the discontinuous nature of the pixel values of adjacent regions and uses derivatives to detect edge points.

The conventional segmentation method will produce different problems in practical applications. In the edge-based seg- mentation method, the segmentation result may face problems such as no edge, severe noise, and excessive smooth boundary; in the threshold-based segmentation method, if the threshold is not set properly, over-segmentation or under-segmentation will occur on the segmentation edge, resulting in false edges or missing edges; in the region-based method, there may be problems such as the size of the segmented region does not match the actual object[7]. In addition, image processing technology based on conventional methods is difficult to meet the requirements of practical applications in terms of segmentation accuracy and segmentation efficiency.

*B. Segmentation methods based on deep learning technology*

Deep learning[8] is a branch of machine learning and a research hotspot in the field of machine learning in the past decade. Deep learning is a method of modeling the information hidden in high-levels by using a multi-layer neural network structure. The main idea of image semantic segmentation technology based on deep learning is to directly input a large amount of original image data to the deep network without artificial design features, and perform complex processing on the image data according to the designed deep network algorithm to obtain high-level abstract features. The output is no longer a simple classification or target location, but a segmented image with the same resolution as the input image with a pixel category label.

As the most successful deep learning model in the field of computer vision, Convolutional Neural Network (CNN) has made breakthroughs in recent years. Since then, deep learning algorithms based on

CNN have continuously made breakthroughs in large-scale competitions in the field of image classification and recognition, and have been widely used in image classification, speech recognition, machine translation and other fields. Its recognition accuracy has even exceeded the accuracy of manual recognition in some areas. Therefore, designing a deep learning model to deal with semantic seg- mentation has great potential.

In 2015, Long et al. proposed a Full Convolutional Network (FCN)[9], which uses a CNN structure with dense predictive capabilities without fully connected layers, which promotes the rapid development of image semantic segmentation. This model allows images of any size to generate segmented images, and it is also much faster than the image block classification method[10]. However, since this method restores the segmentation map with the same size as the original image, the feature map obtained by transposed convolution and upsampling[11] is relatively sparse, which will cause the segmentation result to be less refined, and the method does not take into account the usefulness of global context information. In order to solve this problem, an encoder-decoder structure has been proposed. The encoder down-samples the input image to generate a feature map with lower resolution, which can efficiently classify, and the decoder gradually restores the target details and spatial dimensions. There is usually a shortcut connection between the encoder and the decoder, so it can help the decoder to better repair the details of the target. U-Net is a typical structure in this method[12], it has good performance and simple structure.

Since U-Net was proposed, it has been favored by re- searchers in medical image segmentation. The first network model adopted by most researchers for medical image segmentation is U-Net, and many subsequent models are improved on the basis of U-Net[13,14]. For example, Poudel et al.[15] combined a like U-Net network structure with a cyclic network unit GRU, and proposed RFCN. This network model used the spatial dependence between 2D slices to improve the segmentation of the left ventricular inner and outer membranes of the apex. In addition, there are also some methods that focus on the development of new loss functions for medical image segmentation. Common examples are dice loss[16] and focal loss[17]. Although they can alleviate the class imbalance problem in medical image segmentation tasks and improve segmentation accuracy, but they can't clearly extract precise boundaries. In [18], Kervadec proposed a boundary loss, which explicitly forces the boundary of the segmentation result to be aligned with the boundary of the ground-truth.

In the process of semantic segmentation, the analysis of image semantic scene is extremely critical.

However, most of the early architectures were based on FCN, which did not introduce enough context information and global information under different receptive fields, which could easily lead to wrong segmentation results. In view of the problem that the same object has different scales on different images, the network using multi-scale feature fusion can not only segment large-scale objects, but also segment small-scale objects. Inspired by the SPP module[19], Chen et al.[20] used the spatial pyramid pooling module (ASPP) combined with atrous convolution in DeepLab v2. This module uses atrous convolutions with different sampling rates as parallel branches to extract features of different scales to achieve multi- scale feature fusion. Zhao et al.[21] proposed PSPNet. PSPNet includes a module with hierarchical global priority and different scale information between different sub-regions, called the pyramid pooling module. It makes full use of the prior knowledge of the hierarchical global feature to understand different scenarios, and aggregates the information of different regions to obtain the content of the global context. At the same time, PSPNet also proposed an optimization strategy for moderate supervision loss, which performed well on multiple data sets. It can be said that PSPNet uses both local and global information to achieve the integration of information at different scales and achieves good performance.

The network architecture of semantic segmentation started from FCN [9], and then continued to incorporate effective techniques such as conditional random field (CRF)[22], atrous convolution and spatial pyramid pooling (ASPP)[20] and continuously improve the speed and accuracy of semantic segmentation.

*C.* Deep reinforcement learning

Deep reinforcement learning (DRL) is the product of the combination of deep learning and reinforcement learning. It integrates deep learning's strong understanding of perception problems and the decision-making ability of reinforcement learning to realize end-to-end learning. The emergence of deep reinforcement learning makes reinforcement learning technology truly practical and can solve complex problems in real-world scenarios.

In 2015, Googles DeepMind team developed AlphaGo[23] based on DRL, which pushed DRL to a new hot spot and height, becoming a new milestone in the history of artificial intelligence, and it was widely used in other fields. For example, to optimize the trajectory in unmanned driving[24], apply it to tasks such as text summarization, question and answer[25] and machine translation[26] in NLP, and apply it to automated medical diagnosis and other fields.

Conventional image segmentation work based on deep learning technology largely ignores the dynamic exploration between successive interactions, which greatly reduces the efficiency of segmentation. In our work, we use the agent-based DRL process to draw the segmentation targets contour. By modeling the dynamic process of iteratively updated inter-active image segmentation into a Markov decision process, we have achieved the precise segmentation of the target boundary. This method considers the segmentation update sequence as a whole, makes full use of the correlation and dynamics of continuous interaction, and greatly improves the efficiency and accuracy of segmentation.

## III METHODOLOGY

### A. Data Preprocessing

Our entire dataset contains two-dimensional cardiac images acquired through magnetic resonance imaging (MRI) with non-uniform sizes. For each image, the left ventricle has been outlined by an experienced cardiologist and an associated binary image with the same size has been generated as the contour image, where pixels located on the contour of the left ventricle marked by the cardiologist have value 1 while the remaining pixels have value 0.

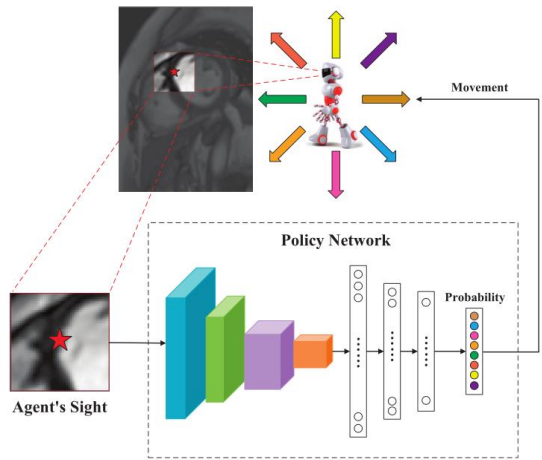

Fig. 1. The overall framework of left ventricle contouring based on reinforcement learning.

We consider each contour of the left ventricle as a sequential point set, which are counter-clockwise sorted without loss of generality. In the original dataset, the contours marked by the cardiologists are commonly discontinuous "trajectories", i.e., two consecutive points on a contour probably fall outside the neighbourhood (a 3 × 3 mask centered by one pixel) of one another. In order to cater for the reinforcement learning setting defined in the next subsection, we have to first refine the original contours by interpolating sufficient extra points between two consecutive ones that are non-adjacent to each other to meet the continuity requirement, i.e., any point on a contour always have neighbours located within its 3×3 mask. After thoroughly inspecting all the contour images, we find that points on the original contours are densely located in vertical direction (i.e., two consecutive points are always located on two adjacent rows) while scarcely located in horizontal direction (i.e., two consecutive points are rarely located on two adjacent columns). Therefore, we can simply scan every two points on a contour and place extra points on every pixel on the row spacing between the two points.

*B.* Problem Definition

Instead of supervised-learning-based image segmentation to highlight the left ventricle, we focus on human-like contouring that mimics how a cardiologist outlines the left ventricle in a cardiac image. Since such a contour drawing process is simply moving a paintbrush along a specific trajectory, it can be analogized to a path finding problem, where a robot explores the unseen world and tries to find its way along the object boundary. In this sense, contouring the left ventricle in a cardiac image can be translated into a reinforcement learning problem, where an agent seeks to make the best decision, i.e., a single-step movement, by interacting with an environment so as to travel as closely as possible to the true contour. In the following part, the key elements of reinforcement learning framework for left ventricle contouring are formally defined.

  *1)* Environment: An environment in left ventricle contouring can be defined as any cardiac image and its associated contour image. Here the whole cardiac image and the true contour of the left ventricle are unseen by the agent. Every time the agent interacts with such an environment by taking a single action, a reward is given by the environment as the feedback to indicate how good that action was, based on which the agent further updates its policy for a better decision, i.e., moving closer to the true contour.

  *2)* State: In the reinforcement learning setting, a state is what the agent can observe, base on which it choose an appropriate action. In this paper, we specify that, being located at any pixel in a cardiac image, the agent can only myopically observe its surrounding sight rather than the whole image. This is an analogy to a path-finding robot with limited eyesight. We define the states as the N×N square mask centered by the pixel where agent is located. One advantage of such setting is that the policy network can be

designed with a uniform structure since the input size is always N×N。

*3)* Action: Since we analogize left ventricle contouring in a cardiac image to path finding in an unseen world, the agent's one-step movement can be defined as the action. Here we specify that at each step, the agent takes an one-pixel move towards one of its eight outbound direction and moves to one of its eight neighboring pixels. Thus, the action a with such definition is discrete and can be numbered with $a \in \{1, ..., 8\}$. Then with the agent's surrounding sight defined as state, Markovian property holds for state transition at each step, i.e., the next state depends only on the current one and the action taken by the agent.

*4)* Reward and Total Return: At each step, the environment gives a reward back to the agent based on its action. Since it is more desirable for the agent to stay as close as possible to the true contour while travelling, reward (denoted by rt for step t) can be defined as the negative distance between the agent's current position (denoted by pt) and the corresponding position on the true contour (denoted by p′t), i.e., $r_t = |p_t - p'_t|$. Therefore, the total return can be defined as the discounted sum of negative distance over the entire episode, i.e.,

$$R = -\sum_{t=1}^{N} \gamma^{t-1} |p_t - p'_t|, \quad (1)$$

where $\gamma \in (0, 1]$ and $T$ denote the discount factor and number of steps in one episode (which differs in different cardiac images).

*5)Episode Termination:* In reinforcement learning frame-work, episode termination indicator is important as well. Normally, an episode can be terminated once the agent has finished the entire steps of actions, the total number of which is equal to the length of the true contour. However, an exception is that an the agent might have moved illegitimately outside the cardiac image before the episode ends normally, which is a common case at the early stage of model training when the agent is not sufficiently "smart" to follow the true contour. In this case, we specify that the episode is terminated in advance and a reward of an extremely small negative value ( -400 throughput the experiments) in this paper) is given to the agent as a penalty for the illegal movement.

## C. Policy Optimization

The previous definition for the elements of reinforcement learning implies a stochastic policy for left ventricle contouring in a cardiac image, i.e., the agent chooses its movement according to a probability distribution, which is output by the policy network with the agent's surrounding sight as input. We resort to proximal policy optimization (PPO) as the training algorithm: a policy-based algorithm which aims at iteratively improving the policy with a relatively small update []. With the PPO algorithm, the policy network is iteratively updated to maximize a clipped surrogate objective function defined as

$$L(\theta) = \mathbb{E}_{s,a \sim \pi_{\theta_k}}[\min(\frac{\pi_\theta(a|s)}{\pi_{\theta_k}(a|s)} A^{\pi_{\theta_k}}(s,a), \\ \mathrm{CLIP}(\frac{\pi_\theta(a|s)}{\pi_{\theta_k}(a|s)}, 1-\epsilon, 1+\epsilon) A^{\pi_{\theta_k}}(s,a))] \quad (2)$$

for iteration $k$, where $\pi_\theta$ denote the policy under parameter $\theta$, $A^\pi(s,a)$ refers to the advantage of action $a$ for state $s$ under policy $\pi$, $\epsilon$ is a hyper-parameter that ranges from 0 to 1 to control the deviation from the current policy $\pi_{\theta_k}$ to the updated one $\pi_\theta$ and the clip function $\mathrm{CLIP}(.,l,u)$ is defined as

$$\mathrm{CLIP}(x,l,u) = \begin{cases} l, x < l \\ x, l \leq x \leq u \\ u, x > u. \end{cases} \quad (3)$$

The PPO algorithm for left ventricle contouring is summarized in Algorithm 1, where Ti refers to the total number of steps in episode trajectory τi, i.e., the contour length. Here the stochastic policy for choosing a movement is trained in an on-policy manner following the "actor-critic" framework in policy-based algorithms. In each iteration, one episode is run according to the up-to-date policy for each training image such that a batch of experience trajectories are collected as training samples, base on which the advantage is estimated. Even with limited training dataset, diversity of the collected episode trajectories can still be guaranteed by randomly choosing the agent's starting point on a contour in each iteration. Then the parameters of the policy network θ and value network φ are alternately updated to maximize the surrogate objective function in (2) ("actor" update) and minimize the mean-square-error between the estimated reward-to-go and the one predicted by the value network ("critic" update) via stochastic gradient ascent/descent.

## D. Generating Landing Spot

Following Algorithm 1, the agent's policy can be iteratively improved such that it is hopefully able to move close to the true contour after rounds of training. During training, the agent's starting point is randomly chosen on a true contour. However, this is unrealistic for testing stage since the true contour of an unseen cardiac image is inaccessible. Therefore, where the agent should start to move on an unseen cardiac image still needs to be figured out. Apparently, the agent's landing spot should be as close to the true contour as possible. Otherwise, if the landing spot is excessively far away from the true contour, it is difficult for the agent to be back on track since its

---

**Algorithm 1** PPO for Left Ventricle Contouring

1: Initial the parameters of policy network $\theta_0$ and value network $\phi_0$;
2: **for** $k = 0, 1, \ldots$ **do**
3:     For each training image, randomly select a position on its contour as the agent's starting point and run one episode by carrying out policy $\pi_{\theta_k}$. A set of episode trajectories $\mathcal{D}_k = \{\tau_i\}$ is then collected for the entire training images;
4:     Estimate advantage for each step based on the current value network:

$$A(s_t, a_t) = \sum_{t'=t}^{T} \gamma^{t'-t} r_{t'} - V_{\phi_k}(s_t);$$

5:     Update the parameters of the policy network to maximize the mean surrogate objective defined in (2):

$$\theta_{k+1} = \arg\max_{\theta} \frac{1}{|\mathcal{D}_k|} \sum_{\tau_i \in \mathcal{D}_k} \frac{1}{T_i} \sum_{t=0}^{T_i} L(\theta);$$

6:     Update the parameters of the value network to minimize the mean squared error between the predicted and estimated value functions:

$$\phi_{k+1} = \arg\min_{\phi} \frac{1}{|\mathcal{D}_k|} \sum_{\tau_i \in \mathcal{D}_k} \frac{1}{T_i} \sum_{t=0}^{T_i} (V_{\phi_k}(s_t) - \sum_{t'=t}^{T} \gamma^{t'-t} r_{t'})^2.$$

7: **end for**

---

surrounding sight in this case could be significantly different from what it observed during training.

Intuitively, a deep-learning model can be trained to generate the coordinate of the agent's appropriate landing spot. After inspecting all the cardiac images, we found that part of true contours always falls into the upper-right quadrant without exception. Therefore, we only focus on the upper-right sub- image with uniform size 100×80 for each cardiac image. The landing spot generator then takes a sub-image as input and is expected to output a coordinate that is as close to any point on the true contour as possible. This can be done through training a supervised-learning model with the loss function defined as the mean minimum distance between the output

coordinate and that of any point on the true contour:

$$L(\psi) = \frac{1}{B} \sum_{i=1}^{B} \min(d_1, ..., d_{T_i}), \quad (4)$$

where

$$d_t = |G_\psi(x) - p'_t| \quad (5)$$

refers to the distance between the output coordinate and the t-th point $p'_t$ on the true contour of a sub-image, G$\psi$(x) refers to the landing spot generator with $\psi$ as parameters and sub-image x as input, Ti denotes the total length of the true contour that falls into the i-th sub-image and B denotes the batch size. In model training, due to the small size of our training set (32 as descried in Section IV -A), the entire training set instead of a batch (a small portion) can be fed into the model in each iteration of parameter update. Therefore, we resort to standard gradient descent to training the model instead of the stochastic gradient descent: instead of being predefined as a static hyper- parameter, the learning rate is optimized through line search, e.g., the secant method. Specifically, in iteration *k* of model training, the optimal learning rate $\lambda_k$ is determined by solving

$$\lambda_k = \arg \max_{\lambda > 0} L(\psi_k - \lambda \nabla L(\psi_k)). \quad (6)$$

The algorithm for training the landing spot generator is summarized in Algorithm 2. Since only a sub-image with smaller size is required as the input and a small portion of a true contour is involved in the loss function, the landing spot generator adapts to different sizes of contour images and the training efficiency can be significantly enhanced even with limited training dataset.

---

**Algorithm 2** Training Landing Spot Generator
1: Initial the parameters of the landing spot generator $\psi_0$;
2: Crop each contour image in the training set to an upper-right sub-image with uniform size;
3: **for** $k = 0, 1, ....$ **do**
4:     Compute the gradient of the loss function in (4) $\nabla L(\psi_k)$ at $\psi_k$;
5:     Find the optimal learning rate $\lambda_k$ by solving (6) through line search;
6:     Update the parameters of the landing spot generator $\psi_{k+1} = \psi_k - \lambda_k \nabla L(\psi_k)$.
7: **end for**

---

The output coordinate of the landing spot generator is not necessarily an integer one, which is probably the case. Therefore, we simply round the output coordinate to its nearest one in the sub-image. Moreover, the output coordinate still has to be translated back to that in the original (full- sized) cardiac image. This can be done by simply leaving the vertical coordinate unchanged while changing the horizontal coordinate *x* following

$$x \Leftarrow x + (N_x - 80), \qquad (7)$$

where Nx refers to the horizontal size of the original cardiac image.

## VI EXPERIMENT

### A. Dataset

Our dataset is provided by the radiology department of a "triple-A-level" hospital in Beijing, China. It contains 100 two-dimensional MRI cardiac images in total, which was collected from 60 patients, and has non-uniform sizes of $208 \times 138$, $208 \times 150$, $208 \times 162$, $208 \times 168$, $208 \times 186$ and $208 \times 210$, respectively. Each cardiac image is associated with a contour image created by an experienced cardiologist as the ground truth. Throughput the experiment, 32 images are selected as the training set, 8 as the validation set while the remaining 60 as the testing set.

### B Model Design

In the experiment, we assume that the agent can only observe its surrounding sight within a $21 \times 21$ patch, which serves as the input of the policy network. The network design of the policy network is given in Table I, which contains two convolution-pooling layers followed by three FC layers. All the layers except the last one adopts "Relu" the activation function. Since the policy network outputs a discrete probability distribution, which corresponds to the probability of moving towards one of the eight outbound direction with one-pixel step, the policy network ends with a "softmax" that outputs eight probabilities.

TABLE I
POLICY NETWORK STRUCTURE

| Layer | Filter size | stride |
|---|---|---|
| Conv2D-16 | $5 \times 5$ | $1 \times 1$ |
| Max_pooling2D | $2 \times 2$ | $2 \times 2$ |
| Conv2D-64 | $3 \times 3$ | $1 \times 1$ |
| Max_pooling2D | $2 \times 2$ | $2 \times 2$ |
| FC-256 | | |
| FC-64 | | |
| FC-8 | | |

Similarly, the value network also takes the agent's surrounding sight as input and outputs the prediction value of the surrounding sight to evaluate the state. The value network is designed following the structure shown in Table II, where comprises two convolution-pooling layers and three FC layers. "Relu" still works as the activation function for all the layers except the last one.

TABLE II
VALUE NETWORK STRUCTURE

| Layer | Filter size | stride |
|---|---|---|
| Conv2D-16 | $5 \times 5$ | $1 \times 1$ |
| Max_pooling2D | $2 \times 2$ | $2 \times 2$ |
| Conv2D-64 | $3 \times 3$ | $1 \times 1$ |
| Max_pooling2D | $2 \times 2$ | $2 \times 2$ |
| FC-256 | | |
| FC-64 | | |
| FC-1 | | |

Structure of the landing spot generator described in Section III-D is shown in Table III. The landing spot generator takes a sub-image with uniform size $100 \times 80$ as input and output the predicted coordinate of the agent's landing spot through three convolution-pooling layers followed by three fully connected (FC) layers. "Relu" is selected as the activation function for all the layers except the last one.

*C Landing Spot*

In this subsection, we evaluate the performance of the proposed landing spot generator. Due to the limited training set of 32 cardiac images, data augmentation is first carried out by flipping each sub-image and its associated contour up-side-down such that the total number of training data is doubled to 64. The training process with 1200 iterations is shown is Fig.2. Although the entire training set instead of only a batch is fed into the model, the decreasing trend in the training curve is still not stable due to the non-convexity of the loss function defined in (4). However, Algorithm 2 still works well for both training and validation sets, i.e., it converges after a few iterations of parameter update through gradient descent with the optimal learning rate and generalization ability of the model improves fast.

TABLE III
STRUCTURE OF LANDING SPOT GENERATOR

| Layer | Filter size | stride |
|---|---|---|
| Conv2D-32 | $5 \times 5$ | $1 \times 1$ |
| Max_pooling2D | $2 \times 2$ | $2 \times 2$ |
| Conv2D-64 | $5 \times 5$ | $1 \times 1$ |
| Max_pooling2D | $2 \times 2$ | $2 \times 2$ |
| Conv2D-64 | $5 \times 5$ | $1 \times 1$ |
| Max_pooling2D | $2 \times 2$ | $2 \times 2$ |
| FC-1024 | | |
| FC-64 | | |
| FC-2 | | |

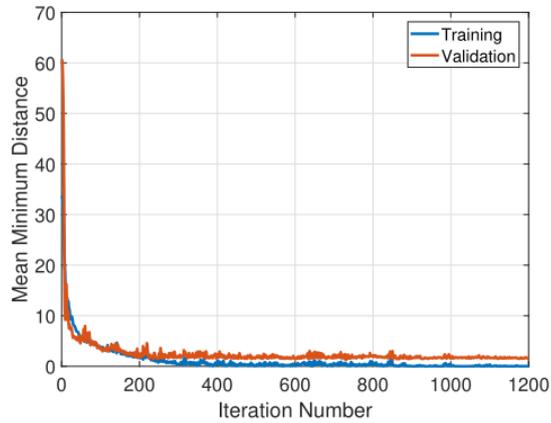

Fig. 2. Training process of the landing spot generator.

For eight cardiac images from the testing set, Fig. 3 demonstrates the agent's landing spots generated by the model specified in Table III, where the red outlines are the ground- truth contours and the blue crosses are the generated landing spots. We can see that the the proposed landing spot generator works well after sufficient training. The coordinate of the point it generates is always sufficiently close to the true contour, which makes the generated landing spot a very good start for the following steps in left ventricle contouring.

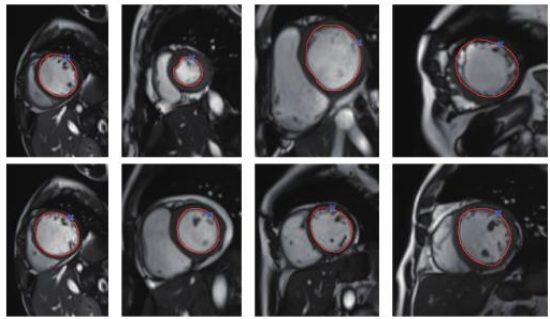

Fig. 3. Landing spots generated by Algorithm 2 for 8 randomly selected cardiac images, where the red the contours are the ground-truth ones and the blue crosses are the generated landing spot.

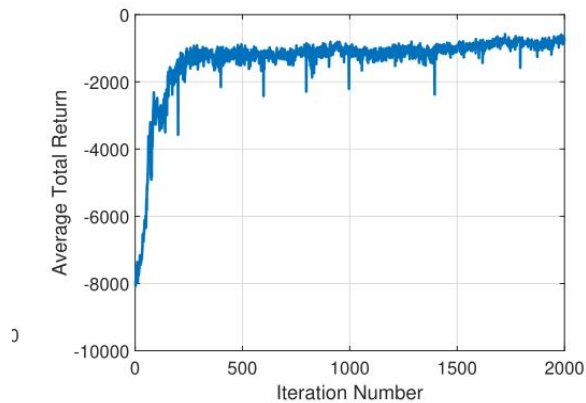

Fig. 4. Training process of the reinforcement learning model.

### D. Contouring Performance

Starting from the landing spot, the agent then sequentially makes its movement decision by following the policy network trained via Algorithm 1 to draw the contour of the left ventricle. Throughout the experiments, the agent's eyesight, i.e., the input of the policy network, is fixed at a $21 \times 21$ square patch. Moreover, during the testing stage when the agent draws the contour on an unseen cardiac image from the testing set, episode termination indicator is still necessary to be designed since the ground-truth contour is never known by the agent. Therefore, we specify that an episode is terminated if the agent have moved into the $3 \times 3$ square patch centered by its first five positions on the contour, which indicates that the agent has finished contouring and returned to the vicinity of the landing spot (e.g., a small area within which it started) since the contour of a left ventricle is normally with a shape of closed circle.

During model training, eight positions on the true contour are randomly selected as the agent's starting points for each cardiac image, which leads to a total of 256 experience trajectories in each iteration for the entire 32 training images. In each iteration, gradient-descent-based parameter update for policy and value networks (step 5 and 6 in Algorithm 1) is alternately carried out with each for 20 times and the learning rate is exponentially decreased during training. To improve training efficiency, we conduct min-max normalization for the reward and z-score standardization for the advantage. Moreover, the network parameters are actually stored in each iteration only when the updated parameters improve the aver-age total return achieved with the validation set.

The training process with 2000 iterations is demonstrated is Fig. 4. In general, Algorithm 1 is able to efficiently converge. However, it is still noticeable that the training process could be temporarily "trapped" in local optimums and the total return even drops in the first 200 iterations. This is a normal case in deep reinforcement learning tasks due to the randomly sampled episode trajectories. Moreover, it can be found that the total return reaches around −570 after convergence. According to the contour images in the training set, the length of a contour amounts to 240 on average, which leads to a per-step deviation of less than 2.4 pixels on average away from the true contour. Considering the sizes of the cardiac images, this is a considerably convincing result in practice due to inevitable uncertainty and randomness in object contouring, i.e., one human expert cannot exactly duplicate a contour drawn by another one (even him-or-herself) for a same object.

We consider two metrics for contouring performance evaluation: Dice score and Hausdorff distance. Dice score is a typical performance indicator for image segmentation and is defined as

$$Dice(\mathcal{S}_p, \mathcal{S}_g) = \frac{2|\mathcal{S}_p \cap \mathcal{S}_g|}{|\mathcal{S}_p| + |\mathcal{S}_g|}, \tag{8}$$

where Sp and Sg denote the point (pixel) sets of the predicted and ground-truth segmentations, respectively, and |.| represents the cardinality of a set, i.e., the total number of pixels in the segment for this case. Hausdorff distance is an important metric for quantitative evaluation the distance between two sets of points and thus can be defined as a performance indicator to evaluate the similarity between two contours. It is defined as

$$d_H(\mathcal{G}, \mathcal{P}) = \max \left\{ \sup_{g \in \mathcal{G}} \inf_{p \in \mathcal{P}} d(\mathcal{G}, \mathcal{P}), \sup_{p \in \mathcal{P}} \inf_{g \in \mathcal{G}} d(\mathcal{G}, \mathcal{P}) \right\}, \tag{9}$$

where G and P are the ground-truth and predicted contours, respectively, and d(g, p) represents the Euclidean distance between two points g and p.

With only 32 cardiac images as the training set, contouring performance is evaluated with a testing set that contains 60 images. The visual result for contouring performance evaluation with eight randomly selected testing images is shown in Fig. 5, where the corresponding segmentation results are provided as well[1]. Although it is difficult to ensure exactly full consistency, the contours of the left ventricle generated via the proposed approach still maintain high resemblance to the ground-truth ones. As previously discussed, these results are reasonably convincing because an radiological expert can never exactly duplicate the contour drawn previously even for the same cardiac image, which is analogous to the fact that one can never exactly duplicate his/her signature. The segmentation results shown in the last two columns of Fig. 5 also illustrate competitive performance of the proposed approach in terms of object segmentation.

We compare the proposed approach with the existing U-net model, which is intended for medical image segmentation based on supervised learning. The result of contouring performance comparison with the same training set of 32 images is summarized in Table IV. Here data augmentation, which triples the train set to 96 images through resizing/cropping to uniform size $208 \times 162$ with horizontal and vertical flipping, is optionally conducted for U-net model training.

According to the numerical results in Table IV, the proposed reinforcement-learning-based approach for left ventricle contouring achieves 92.6% in average Dice score and 5.784 in average Hausdorff distance, which is a satisfactory result, especially considering our limited training set of only 32 images. Table IV also shows that in general, the proposed approach outperforms the U-net model with limited training set. An exception is that the proposed approach falls a little behind (while is still comparable to) the U-net model with data augmentation in average Dice score. Especially, for average Hausdorff distance, the proposed approach shows a significant advantage over the U-net model, even with data augmentation. This can be well explained by how the proposed reinforcement-learning-based approach works, i.e., it focuses on drawing the contour of an object instead of segmentation. Fig. 6 illustrates the visual result of segmentation performance comparison on randomly selected images. The last two columns show that with limited training set, the segmentation performance of the U-net model (even with data augmentation) is less satisfactory on cardiac images with irregular plaques on the left ventricles, where unexpected "grooves" and "holes" appear in the segmentation. In contrast, similar with Fig. 5, left ventricles identified by the proposed approach are still highly consistent with the ground-truth ones. This is still owed to the fact that the proposed reinforcement-learning-based approach aims at drawing contours as close to the ground-truth ones as possible. With well trained policy network in the proposed approach, those irregular plaques never distract the agent in its movement decision making.

## V. CONCLUSION

In this paper, we propose a new iterative and refined interactive segmentation method for medical images based on agent reinforcement learning. This method explores an interactive policy that iteratively combines the hints obtained by the agent to improve image segmentation performance. We focus on the segmentation boundary problem of the target, and model the dynamic process of drawing the target contour in a certain order as a Markov Decision Process (MDP) based on a deep reinforcement learning method. In the dynamic process of continuous interaction between the agent and the image, the agent tracks the boundary point by point in order within a limited length range until the targets contour is completely drawn. In this process, the agent continuously adjusts its actions according to the rewards received, thereby achieving a more precise and finer segmentation boundary. At the same time, in each iteration process, the agent can quickly improve the segmentation performance. Experimental results show that our method has a better segmentation effect on the left ventricle in a small number of medical image data sets, especially in terms of segmentation boundaries, this method is better than existing methods. In addition, our method can be easily applied to other challenging segmentation tasks proposed in various practical applications.

TABLE IV
CONTOURING PERFORMANCE COMPARISON

| Metrics | Average Dice Score ($\pm$Standard Deviation) | Average Hausdorff Distance ($\pm$Standard Deviation) |
|---|---|---|
| U-net without data augmentation | 0.896 ($\pm$0.075) | 14.158 ($\pm$13.386) |
| U-net with data augmentation | 0.935 ($\pm$0.052) | 11.966 ($\pm$11.784) |
| Ours | 0.926 ($\pm$0.043) | 5.784 ($\pm$1.713) |

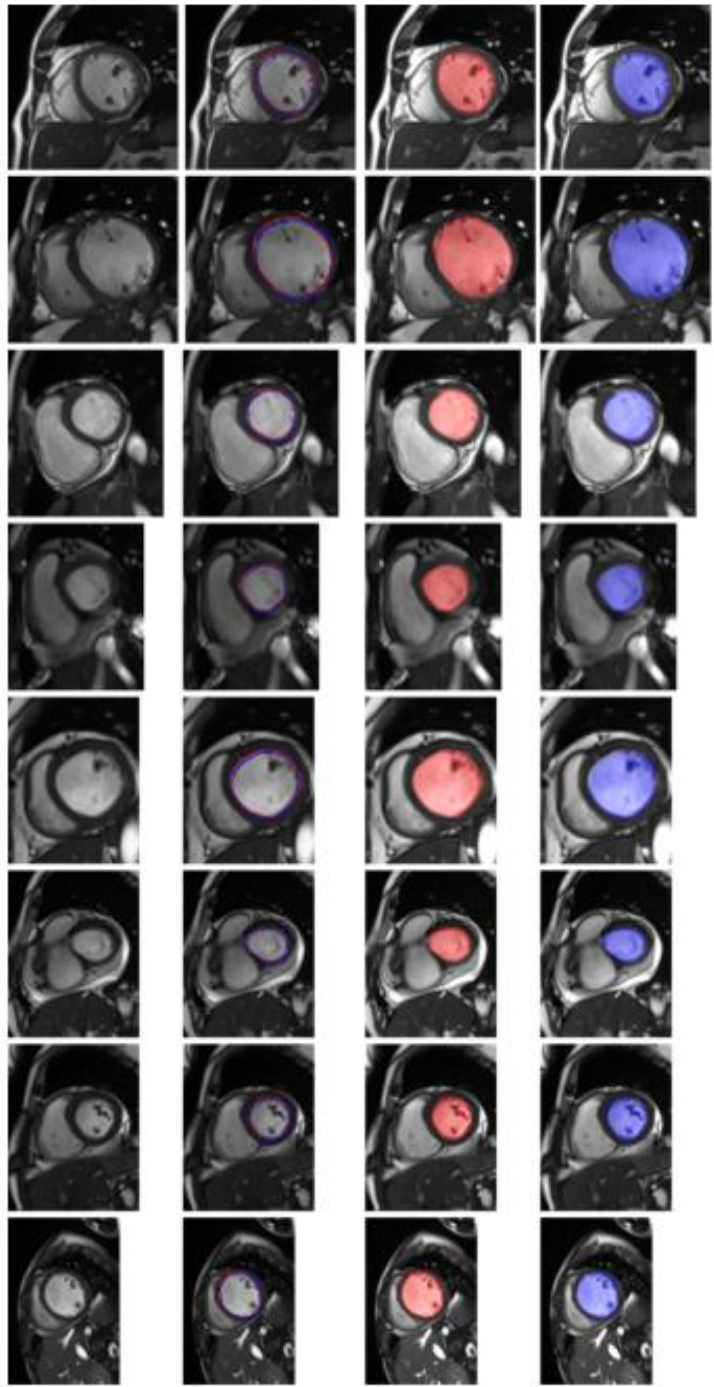

Fig. 5. Segmentation results of the left ventricle in 8 sample images based on the deep reinforcement learning method. From left to right, the example-original MR image, ground truth contour (red) and predicted contour (blue), ground truth patch, predicted segmentation results are shown respectively.

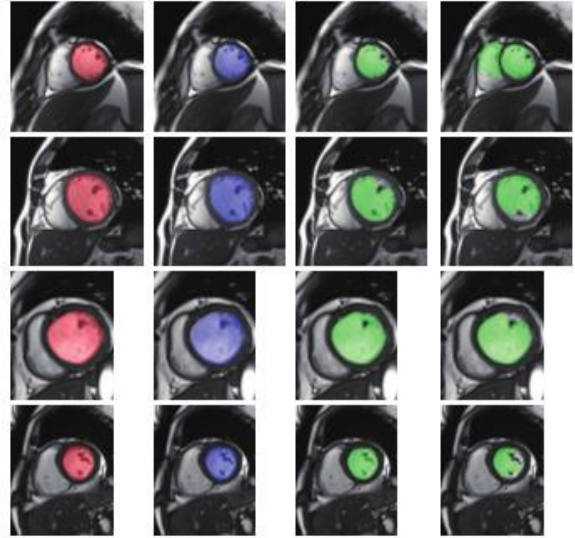

Fig. 6. Visual comparison for segmentation performance with four sample images. From left to right are the ground truth, the proposed approach, U-net without data augmentation and U-net with data augmentation.